\documentclass[conference]{IEEEtran}
\IEEEoverridecommandlockouts
\usepackage{cite,balance}
\usepackage{amsmath,amssymb,amsfonts}
\usepackage{amsthm}
\usepackage{algorithmic}
\usepackage{graphicx}
\usepackage{textcomp}
\usepackage{xcolor}
\usepackage{mathtools}
\def\BibTeX{{\rm B\kern-.05em{\sc i\kern-.025em b}\kern-.08em
    T\kern-.1667em\lower.7ex\hbox{E}\kern-.125emX}}

\newcommand\numberthis{\addtocounter{equation}{1}\tag{\theequation}}

\usepackage{amsmath}

\newtheorem{theorem}{Theorem}[section]

\newtheorem{lemma}{Lemma}[section]

\newtheorem{remark}{Remark}[section]

\begin{document}

\title{Coding-Enforced Resilient and Secure Aggregation for Hierarchical Federated Learning \\
\thanks{This work was funded in part by the Swedish Foundation for Strategic Research (FUS21-0026) and in part by the Swedish Innovation Agency (Vinnova) through the SweWIN Center (2023-00572).}
}

\author{\IEEEauthorblockN{Shudi Weng, Ming Xiao, Mikael Skoglund}
\IEEEauthorblockA{\textit{Department of Information Science and Engineering} \\
\textit{KTH Royal Institute of Technology}\\
Stockholm, Sweden \\
\{shudiw, mingx, skoglund\}@kth.se
}
}

\maketitle

\begin{abstract}
Hierarchical federated learning (HFL) has emerged as an effective paradigm to enhance link quality between clients and the server. However, ensuring model accuracy while preserving privacy under unreliable communication remains a key challenge in HFL, as the coordination among privacy noise can be randomly disrupted. 
To address this limitation, we propose a robust hierarchical secure aggregation scheme, termed H-SecCoGC, 
which integrates coding strategies to enforce structured aggregation. 
The proposed scheme not only ensures accurate global model construction under varying levels of privacy, but also avoids the partial participation issue, thereby significantly improving robustness, privacy preservation, and learning efficiency.
Both theoretical analyses and experimental results demonstrate the superiority of our scheme under unreliable communication across arbitrarily strong privacy guarantees.
\end{abstract}

\section{Introduction}
Federated learning (FL), as a promising edge training paradigm, enables coordination between a central server and edge clients to collaboratively train a shared model without transmitting the raw datasets. 
To enhance client-to-server link quality, an effective strategy is to use advanced transmission technologies, e.g., relay-assisted hierarchical federated learning (HFL) \cite{lin2022relay,chen2023relay}, which introduces intermediate nodes that mitigates the adverse effects of channel fading in wireless systems and facilitates data exchange between mobile devices and the server. 
Even so, HFL still confronts the issue of unreliable communication, which can lead to unpredictable partial participation and degrading training performance.



\subsubsection{Coding Strategies in Straggler Mitigation}
Partial client participation can reduce the effective step length \cite{weng2025heterogeneity}, slowing down the training speed, and lead to objective inconsistency\cite{wang2021quantized}, making the trained model deviate from the global optimum. 
Among existing solutions, coding-based strategies are particularly appealing, as they offer strong robustness without requiring prior information \cite{li2025adaptive,9838986,10806940,weng2024coded}. 
The most notable coding scheme is gradient coding (GC), initially proposed by Tandon et al. \cite{tandon2017gradient}, and subsequently extended to accommodate various practical needs, such as improving communication efficiency \cite{ye2018communication}, handling heterogeneous computational capabilities \cite{wang2019heterogeneity,jahani2021optimal}, and exploiting temporal structures \cite{krishnan2021sequential}. 
However, these GC schemes rely on dataset replication across clients, and thus, are not suitable for federated learning (FL).
Moreover, the dataset replication leads to substantial communication and computation load. 
To address these issues, \cite{weng2024cooperative,11175173} develops a gradient-sharing GC framework, termed cooperative gradient coding (CoGC), so that each client only trains a single dataset. 
\subsubsection{Privacy-Preserving HFL/FL} 
Another critical challenge in HFL is the privacy leakage of the sensitive information contained in the local models. A large number of studies investigate the use of correlated/uncorrelated Gaussian noises in HFL/FL \cite{9685644, rodio2025optimizingprivacyutilitytradeoffdecentralized}, confirming the effectiveness of the Gaussian mechanism. 
Alternative approaches include the use of coded datasets \cite{li2025adaptive}.
However, these works are constrained by the inherent privacy-utility trade-off. 
In contrast, secure aggregation enables full cancellation of the privacy noises, ensuring accurate global model construction under arbitrary privacy levels \cite{bonawitz2017practical,9929413}. 
However, this advantage heavily relies on stable communication. To address this, our previous work proposes a coding-enforced structured aggregation method in cooperative networks \cite{weng2025codingenforcedrobustsecureaggregation}. 

To achieve robust, accurate global model construction while preserving privacy, we extend the coding-enforced secure aggregation framework \cite{weng2025codingenforcedrobustsecureaggregation} to HFL.

Our main contributions are summarized as follows. 
\begin{itemize}
    \item We propose H-SecCoGC\footnote{This paper mainly addresses the learning and privacy performance, whereas the fundamental limits of communication and secret key rate of the the method has been established in  \cite{zhang2025fundamentallimitshierarchicalsecure,weng2026resilientefficientlinearsecure} accounts for H-SecCoGC.}, a privacy-preserving and straggler-resilient HFL framework based on gradient-sharing and cyclic GC codes that ensures exact global model recovery under arbitrarily strong privacy over intermittent networks. The proposed scheme ensures reliable cancellation of privacy noises at server aggregation through coding structure, thereby avoiding partial participation issues and achieving substantial performance improvement without requiring prior information. 
    \item We investigate the worst-case information leakage under optimal secret keys by formulating a max–min problem, which identifies the optimal keys and the worst-case source distribution of local models.
    \item We present a quantitative analysis of local differential privacy (LDP) across all layers of the proposed secure HFL protocols. The analysis jointly considers unreliable communication, correlations among privacy noise, and the inherent stochasticity of local models.
\end{itemize}


\section{System Model and Preliminaries}
\subsection{Hierarchical Federated Learning}
Consider a distributed system consisting of $K$ clients, each holding local datasets $\mathcal{D}_k$ on it, and $K$ relays, acting as intermediate nodes during communication between clients and the server, coordinated by a central server. The goal is to learn a global model $\boldsymbol{\Theta}\in\mathbb{R}^D$ that minimizes the \textit{global objective function (GOF)} defined below: 
\begin{align}
    \min_{\boldsymbol{\Theta}} \mathcal{L}(\boldsymbol{\Theta})\triangleq \frac{1}{K}\sum_{k=1}^{K} \mathcal{L}_k(\boldsymbol{\Theta}, \mathcal{D}_k),
    \label{eq: goal}
\end{align}
through iterative training and communication. The \textit{local objective function (LOF)} $\mathcal{L}_k(\cdot)$ that measures the accuracy of the model $\boldsymbol{\Theta}$ on local datasets $\mathcal{D}_k$ is defined by the average loss on all training samples in $\mathcal{D}_k$, that is,
\begin{align}
    \mathcal{L}_k(\boldsymbol{\Theta}, \mathcal{D}_k)\triangleq \frac{1}{\lvert \mathcal{D}_k \rvert}\sum_{\xi\in \mathcal{D}_k} l(\boldsymbol{\Theta}, \xi),
    \label{eq: LOF}
\end{align}
where $l(\boldsymbol{\Theta}, \xi)$ is the sample loss of the model $\boldsymbol{\Theta}$ evaluated on a single training sample $\xi$. 

In the $t$-th training round, where $t\in[T]$, all clients are initialized with the latest global model, i.e., $\boldsymbol{\Theta}_k^{t,0}=\boldsymbol{\Theta}^{t-1}$. The local training on client $k$ at the $i$-th iteration is given by   
\begin{align}
    \boldsymbol{\Theta}_k^{t,i}=\boldsymbol{\Theta}_k^{t,i-1}-\eta a_{i} \nabla \mathcal{L}_k(\boldsymbol{\Theta}_k^{t, i-1}, \boldsymbol{\xi}_k^{t, i}), i\in [I]
\end{align}
where $\eta$ is the learning rate, $I$ is the number of local iterations, $a_{i}$ characterizes the type of local solver, $\boldsymbol{\xi}_k^{t, i}$ is the data patch containing $n_p$ data samples extracted from $\mathcal{D}_k$ with $\lvert \boldsymbol{\xi}_k^{t, i} \rvert=n_p$. The gradient $\nabla\mathcal{L}_k(\boldsymbol{\Theta}_k^{t, i-1}, \boldsymbol{\xi}_k^{t, i})$ computed on the data patch $\boldsymbol{\xi}_k^{t, i}$ is defined by the averaged gradients over all data samples $\xi\in\boldsymbol{\xi}_k^{t, i}$. In matrix form, the local model update $ \Delta\boldsymbol{\Theta}_k^{t,I}=\boldsymbol{\Theta}_k^{t,I}-\boldsymbol{\Theta}^{t}$ can be expressed by
\begin{align}
    \Delta\boldsymbol{\Theta}_k^{t,I}=-\eta\boldsymbol{a}\cdot \nabla\boldsymbol{\mathcal{L}}_k^{t,I}, 
    \label{eq: local training}
\end{align}
where $ \boldsymbol{a}=\left[a_{1}, \cdots, a_{I} \right]$ describes how the gradients accumulate over iterations \cite{wangTackling}, and the gradients obtained at all iterations are stored in $\nabla\boldsymbol{\mathcal{L}}_k^{t,I}$, defined as $\nabla\boldsymbol{\mathcal{L}}_k^{t,I}=\left[ \nabla \mathcal{L}_k(\boldsymbol{\Theta}_k^{t, 0}, \boldsymbol{\xi}_k^{t, 1}), \cdots, \nabla \mathcal{L}_k(\boldsymbol{\Theta}_k^{t, I-1}, \boldsymbol{\xi}_k^{t, I}) \right]^\top $. 

After local training, the clients communicate with relays, and the relays forward to the server. With full participation of clients, the server should be able to update $\boldsymbol{\Theta}^t$ as
\begin{align}
\boldsymbol{\Theta}^{t}=\boldsymbol{\Theta}^{t-1}+\frac{1}{K}\sum_{k\in[K]} \Delta\boldsymbol{\Theta}_k^{t,I}.
    \label{eq: sever_agg}
\end{align}

\subsection{Communication Model}
All links (both client-to-relay and relay-to-server communications) are assumed to be either unavailable or perfect, i.e., interference or channel distortions are beyond considerations.
The link $\tau^t_j$, from relay $j$ to the sever, is modeled by $\tau^t_j\sim \mathrm{Ber}(1-p_j)$; the link $\tau^t_{j,k}$, from client $k$ to relay $j$, is modeled by $\tau^t_{j,k}\sim \mathrm{Ber}(1-p_{j, k})$, where $p_j$ and $p_{j,k}$ characterize the channel outages. The $\tau^t_j/\tau^t_{j,k}=1$ implies a successful communication, and $\tau^t_j=0/\tau^t_{j,k}=0$ implies a complete failure.
Additionally, all links are assumed to be statistically independent.

\subsection{$(K,s)$-Cyclic Gradient Codes}
The GC codes are defined from a cyclic coefficient matrix $\boldsymbol{G}$ and a combination matrix $\boldsymbol{C}$ that are pairwise designed. The $\boldsymbol{G}$ is devised such that the all-one vector lies in the span space of any subset of its $K-s$ columns. The rows in $\boldsymbol{C}$ spans the $K-s$ columns in $\boldsymbol{G}$ such that $\boldsymbol{C}\boldsymbol{G}=\mathbf{1}_{f\times K}$, where $f={K \choose s}$, the nonzero pattern of rows in $\boldsymbol{C}$ encompasses all possible $s$-straggler patterns. Such structured codes ensure that the all-one vector is attainable against any $s$ stragglers.

\section{The Proposed Method}\label{sec:H-SecCoGC}
\subsection{Description}

\begin{figure}
    \centering
    \includegraphics[width=0.9\linewidth]{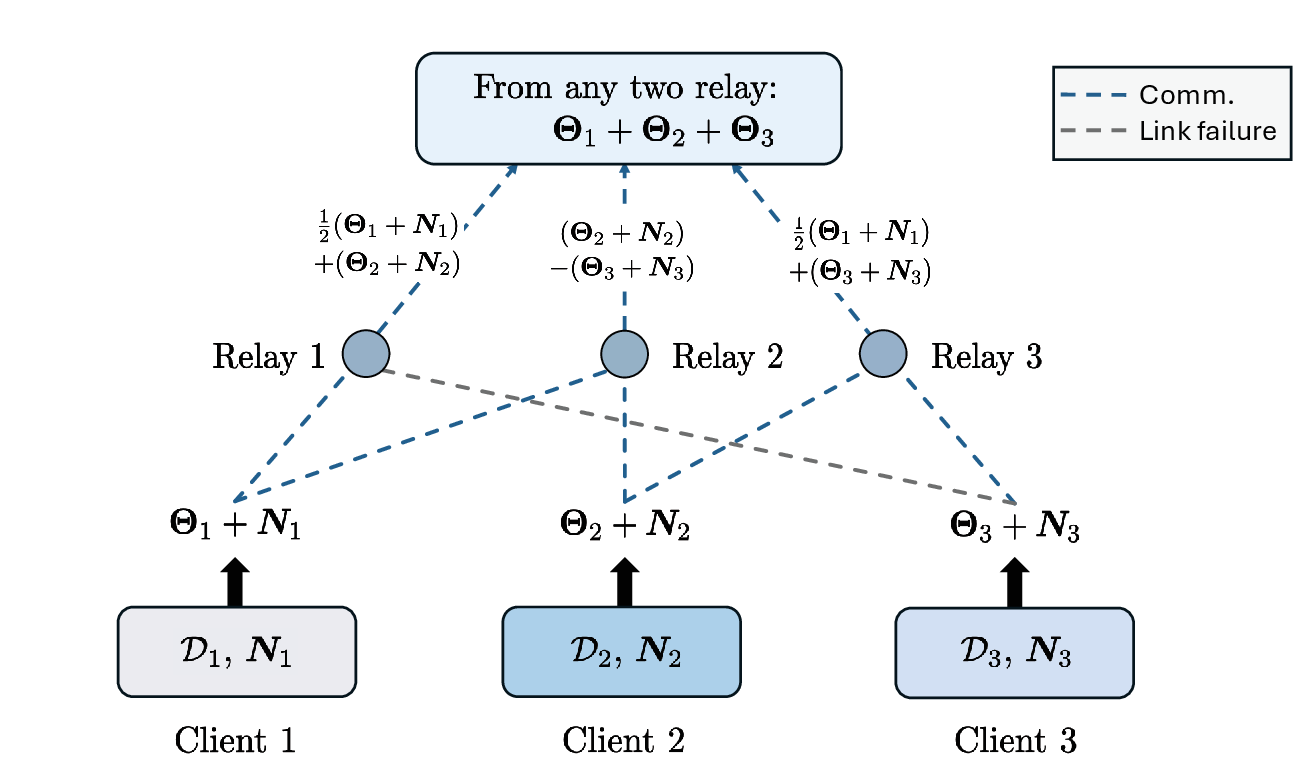}
    \caption{ The proposed robust secure aggregation based on hierarchical CoGC with $K=3$ $s=1$ under unreliable communication condition. }
    \label{fig: H_SecCoGC}
\vspace{-3mm}
\end{figure}
The proposed scheme, hierarchical secure cooperative gradient coding (H-SecCoGC), is illustrated in Fig. \ref{fig: H_SecCoGC}. Suppose that there are $K$ relays. The clients associated with relay $j$ is assigned according to the nonzero pattern of the $j$-th row in coding coefficient matrix $\boldsymbol{G}$. If the $(j,k)$-th element in $\boldsymbol{G}$ is nonzero, i.e., $g_{j,k}\neq 0$, then client $k$ is required to upload its updates to relay $j$. Accordingly, each client $k$ is associated with relays in $\mathcal{R}_k=\{j: g_{j,k}\neq 0\}$, each relay $j$ receives from clients in $\mathcal{W}_j=\{k: g_{j,k}\neq 0\}$.
Notably, there may exist indices $j_1$ and $j_2$ such that the same client $k$ is associated with both relay $j_1$ and $j_2$. 
 Below, we omit the index $I$.
 
Each client stores a secret key $\boldsymbol{N}_k^t$ for secure transmission. The keys are designed such that 
\begin{align}
    \sum_{k\in[K]}\boldsymbol{N}_k^t=0. 
\end{align}
The construction of such secret keys $\boldsymbol{N}_k^t$ is discussed in \cite{weng2025codingenforcedrobustsecureaggregation}.

After local training, each client masks its local model update by adding privacy noise to it, i.e., 
\begin{align}
    \boldsymbol{Y}_k^t=\Delta\boldsymbol{\Theta}_k^t+\boldsymbol{N}_k^t. 
\end{align}
Then, each client $k$ shares its masked local update $\boldsymbol{Y}_k^t$ with relay $j\in\mathcal{R}_k$. 
By transmitting masked local model updates, no other third party, e.g., other clients, relays, or the server, can learn about any individual local update $\Delta\boldsymbol{\Theta}_k^{t,I}$ accurately, thereby protecting local privacy. 

 The transmissions from clients to relays may fail, depending on $\tau^t_{j,k}$. If relay $j$ eventually received messages from its associated clients in $\mathcal{W}_k^t$, i.e., $\mathcal{W}_j^t=\{k: \tau^t_{j,k}=1\}$, then relay $j$ computes an equation according to the coefficient matrix $\boldsymbol{G}$,
 \begin{align}
     \boldsymbol{S}^t_j=\sum_{k\in \mathcal{W}_j^t} g_{j,k} \left(\Delta\boldsymbol{\Theta}_k^t+\boldsymbol{N}_k^t \right), 
     \label{eq: partial_sum}
 \end{align}
and send \eqref{eq: partial_sum} to the central server for further aggregation. 
If a relay fails collect all local model updates from clients in $\mathcal{W}_j$, the computation in \eqref{eq: partial_sum} is referred to as a complete partial sum.

If the server receives at least $K-s$ complete partial sums from all $K$ relays, it can combine these equations to learn the exact global model according to $\boldsymbol{C}$. To perform this, the server first needs to determine which combination row to use by comparing the nonzero patterns of rows in $\boldsymbol{C}$ with the connectivity from relays to the server. For the $f_t$-th row, if for $\forall j: \tau^t_j=0$, the corresponding element $c_{f_t, j}=0$, then the server can use the $f_t$-th row to compute the global model.
\begin{align}
    \frac{1}{K}\sum_{j=1}^K c_{f_t,j}\boldsymbol{S}^t_j
    =\frac{1}{K}\sum_{k\in[K]} \left(\Delta\boldsymbol{\Theta}_k^t+\boldsymbol{N}_k^t\right)
    =\frac{1}{K}\sum_{k\in[K]} \Delta\boldsymbol{\Theta}_k^t. 
\end{align}  
When the server receives fewer than $K-s$ equations, no meaningful results can be reconstructed. That is, the server can either perfectly reconstruct the global model with full client participation, or fail to perform a meaningful aggregation otherwise. For this reason, the system is designed so that client-to-relay and relay-to-server communications repeat until the server can compute the correct global model. Through such an adoption of the coding-based approach and a structured secret key construction, the privacy noise is fully eliminated in successful aggregations while remaining in failed ones.  
In this way, convergence is guaranteed, and the convergence rate matches that of standard FL with perfect communications \cite{li2019convergence}. 

\begin{remark}[Communication Costs]
In wireless settings, each client only broadcasts their masked local model once, and each relay only broadcasts the partial sum once. 
\end{remark}
\begin{remark}[Decoding Complexity]
Each relay needs to decode $s+1$ messages from different clients. The server needs to decode at most $K$ messages from relays. 
\end{remark}

\subsection{Optimal Secret Keys}
What is the maximum information leakage during the gradient-sharing phase when the optimal secret keys are used? 
To answer this question, each secret key is assumed to satisfy a power constraint, $\operatorname{Diag}(\operatorname{Cov}[\boldsymbol{N}_k^t]) \leq \lambda^2 \boldsymbol{I}_D$, 
each local update is assumed to satisfy a total power constraint, i.e., $\operatorname{Tr}(\operatorname{Cov}[\Delta\boldsymbol{\Theta}_k^t]) \leq D\zeta^2$. 
Moreover, the local model updates $\{\Delta\boldsymbol{\Theta}_k^t\}_{k\in[K]}$ are assumed to be mutually independent since local training is performed independently on each client.

Consider the transmission from client $k$ to $m$, $\forall k$, $\forall m$, the following max-min problem is of particular interest.
\begin{align}
    &\adjustlimits\sup_{\Delta\boldsymbol{\Theta}_k^t} \inf_{\boldsymbol{N}_k^t} \mathrm{I}\left(\Delta\boldsymbol{\Theta}_k^t; \tau^t_{m,k}(\Delta\boldsymbol{\Theta}_k^t+\boldsymbol{N}_k^t)\right) \notag\\
    =&\adjustlimits\sup_{\Delta\boldsymbol{\Theta}_k^t} \inf_{\boldsymbol{N}_k^t} (1-p_{m,k})\mathrm{I}\left(\Delta\boldsymbol{\Theta}_k^t; \Delta\boldsymbol{\Theta}_k^t+\boldsymbol{N}_k^t \right) 
    \notag\\
    =&\adjustlimits\sup_{\Delta\boldsymbol{\Theta}_k^t} \inf_{\boldsymbol{N}_k^t} 
    (1-p_{m,k})\left(\mathrm{h}(\Delta\boldsymbol{\Theta}_k^t+\boldsymbol{N}_k^t)
    -\mathrm{h}(\boldsymbol{N}_k^t)\right) \notag\\
    =& (1-p_{m,k})\left(\mathrm{h}\left(\mathcal{N}(\mathbf{0}, (\lambda^2+\zeta^2) \boldsymbol{I}_D)\right)
    -\mathrm{h}\left(\mathcal{N}(\mathbf{0}, \lambda^2 \boldsymbol{I}_D)\right)\right) \notag\\
    =& (1-p_{m,k})\frac{D}{2}\log\left(1+\frac{\zeta^2}{\lambda^2}\right),
\end{align}
where the above result follows \cite[(12)]{959289} and that the differential entropy is maximized by a Gaussian vector whose covariance matrix is diagonal with identical diagonal entries.

\begin{lemma}[Worst Input]
    Any other distribution of the local model updates than the Gaussian vector with a diagonal covariance matrix $\zeta^2\boldsymbol{I}_D$ results in lower privacy leakage. 
\end{lemma}
\begin{lemma}[Optimal Keys]\label{corollary:optimal key}
    The secret key $\boldsymbol{N}_k^t\sim\mathcal{N}(\mathbf{0}, \lambda^2\boldsymbol{I}_D)$ is optimal and results in the smallest privacy leakage under a total power constraint of $D\lambda^2$.
\end{lemma}
In the analysis below, we employ the optimal secret keys for $\forall k$.

\section{Privacy Analysis}
This section evaluates the local differential privacy (LDP) at each layer in H-SecCoGC, which accounts for both the correlation among secret keys and the effects of unreliable communication.
Compared with \cite{weng2025codingenforcedrobustsecureaggregation}, the main distinction is that $(i)$ all local models are considered as Gaussian random variables, instead of a deterministic value, $(ii)$ the relay node does not have any knowledge about the secret keys, and $(iii)$ there is communication from the $k$-th client to the $k$-th relay node. 
In the analysis below, it is assumed that the local models are i.i.d. Gaussian distributed, i.e., $ \Delta\boldsymbol{\Theta}_k^t\sim \mathcal{N}(\boldsymbol{0}, \zeta^2\boldsymbol{I}_D)$. 
This assumption is realistic, since the superposition of multiple independent noise sources tends to produce Gaussian noise according to the law of large numbers.
\begin{lemma}\label{lemma:chi_square bound}
   For $\Delta\boldsymbol{\Theta}_k^t\sim\mathcal{N}(\boldsymbol{0}, \zeta^2\boldsymbol{I}_D)$, the following Chi-squared tail bound holds for $\forall \delta_0$,
\begin{align}
    \mathrm{P_r}\left(\left\lVert \Delta\boldsymbol{\Theta}_k^t \right\rVert\hspace{-1mm}\leq\hspace{-1mm} \zeta\sqrt{D(1+\delta_0)}\right)\hspace{-0.5mm}\geq\hspace{-0.5mm} 1\hspace{-0.5mm}-\hspace{-0.5mm} e^{-\frac{D}{2}(\delta_0-\ln{(1+\delta_0)})}.
    \label{eq: chi_tail}
\end{align}
\end{lemma}

\subsection{Client-level LDP}
The client $k$ sends its masked local model $\boldsymbol{Y}_k^t$ to relays in $ \mathcal{R}_k$. During the transmission, the local model $\Delta\boldsymbol{\Theta}_k^t$ should not be confidently inferred. This leads to two cases: (i) if the communication link is unavailable, the transmitted message remains perfectly private, since no information about client $k$ is revealed; (ii) if $\boldsymbol{Y}_k^t$ is successfully received, the relay should not distinguish $\Delta\boldsymbol{\Theta}_k^t=\boldsymbol{u}$ and $\Delta\boldsymbol{\Theta}_k^t=\boldsymbol{u}'$ precisely. 
\begin{theorem}\label{theo: p2p LDP}
    Given the secret key $\boldsymbol{N}_k^t\sim\mathcal{N}(\mathbf{0}, \lambda^2\boldsymbol{I}_D)$, $R=\zeta\sqrt{D(1+\delta_0)}$, and $\forall \delta_{m, k}^{(1)}\in(0,1]$, $\boldsymbol{Y}_k^t$ from client $k$ to relay $j$ via unreliable link $\mathrm{Ber}(1-p_{j,k})$ is $(\epsilon_{m, k}^{(1)},(1-p_{j,k})\delta_{m, k}^{(1)})$-differentially private w.p. $1\hspace{-0.5mm}-\hspace{-0.5mm} e^{-\frac{D}{2}(\delta_0-\ln{(1+\delta_0)})}$, where
    \begin{align}
        \epsilon_{m, k}^{(1)}= \frac{R}{\lambda}\left[2\log\left( \frac{1.25}{\delta_{m, k}^{(1)}} \right)\right]^{\frac{1}{2}}, 
    \end{align}
    i.e., for any measurable set $\mathcal{S}$, it holds that
    \begin{align*}
         \mathrm{P_r}(\boldsymbol{Y}_k^t\in \mathcal{S} \vert \Delta\boldsymbol{\Theta}_k^t=\boldsymbol{u})
         \leq& e^{\epsilon_{j,k}^{(1)}}\mathrm{P_r}(\boldsymbol{Y}_k^t\in \mathcal{S} \vert \Delta\boldsymbol{\Theta}_k^t=\boldsymbol{u}')\\ 
         &\hspace{10mm}+(1-p_{j,k})\delta_{j,k}^{(1)}.
         \numberthis
    \end{align*}
\end{theorem}
\begin{proof}[Proof Sketch]
The proof begins with Lemma \ref{lemma:chi_square bound} to bound $\Delta\boldsymbol{\Theta}_k^t$ with a probability guarantee and proceeds by applying standard Gaussian LDP.
\end{proof}
\vspace{-0.5em}
\subsection{Relay-level LDP}
If relay $j$ successfully receives messages from clients in $\mathcal{W}_j^t$, it computes the partial sum $\boldsymbol{S}_{j}^t$ and transmits it to the server. 
The server observes the masked partial sum $\boldsymbol{S}_{j}^t$, it should not confidently infer whether client $k$'s contribution is included in $\boldsymbol{S}_{j}^t$. 
The privacy is provided by: \textit{(i)} the link disruption from client $k$ to relay $j$, and \textit{(ii)} the aggregated correlated privacy noises.
Let $\boldsymbol{M}_j^t=\sum_{k\in \mathcal{W}_j^t} g_{j,k} \boldsymbol{N}_k^t $ denote the aggregated privacy noise at relay $j$, and let $\boldsymbol{M}_j^{t, -k}=\sum_{m\in\mathcal{W}_j^t\setminus \{k\}} g_{j,m}\boldsymbol{N}_m^t$ denote the aggregated privacy noise at relay $j$ excluding client $k$'s privacy noise. 
Similarly, let $\boldsymbol{S}_{j}^{-k}\triangleq \sum_{m\in\mathcal{W}_j^t\setminus \{k\}} g_{j,m}(\Delta\boldsymbol{\Theta}_m^t+\boldsymbol{N}_m^t)$ denote the partial sum at relay $j$ excluding client $k$’s contribution. The expressions of variances $\nu_j^t=\mathrm{var}(\boldsymbol{S}^t_j)$ and $\nu_j^{t, -k}=\mathrm{var}(\boldsymbol{M}_j^{t, -k})$ are given in \cite[Appendix G]{weng2025codingenforcedrobustsecureaggregation}. Note that $\nu_j^t$ and $\nu_j^{t, -k}$ are both random variables with respect to intermittent network topology, yet this challenge can be effectively addressed through an appropriate choice of the Bernstein parameter \cite{saha2024privacy}. 

\begin{theorem}\label{theo: relay LDP 1}
Given matrix $\boldsymbol{A}$, which generates the secret keys $ \{\boldsymbol{N}_k^t\}_{k\in[K]}$ as in \cite[(14)]{weng2025codingenforcedrobustsecureaggregation}, the variance of $\boldsymbol{M}_j^t$ is $\nu_j^{t, -k} \boldsymbol{I}_D=\lVert \boldsymbol{\Lambda}_j^{t, -k} \boldsymbol{A}\rVert^2 \boldsymbol{I}_D$,\footnote{The \( k \)-th element of the vector \( \boldsymbol{\Lambda}_j^t\) is defined as
$\boldsymbol{\Lambda}_j^t(k) = \tau^t_{j,k} g_{j,k}$.
Accordingly, its expectation \( \boldsymbol{\Lambda}_j^{\mathrm{E}} \) is given by
$\boldsymbol{\Lambda}_j^{\mathrm{E}}(k) = (1 - p_{j,k}) g_{j,k}$. 
Different from \cite{weng2025codingenforcedrobustsecureaggregation}, in HFL, $\forall K: \tau^t_{k,k}$ is likely to be zero. 
}  and $\Bar{\nu}_j^{-k}=\mathbb{E}[\nu_j^{t, -k}]=\left(\lambda^2\sum_{m=1}^K(1-p_{j,m})g_{j,m}^2+\boldsymbol{\Lambda}_j^{\mathrm{E},-k}\boldsymbol{A}\boldsymbol{A}^\top(\boldsymbol{\Lambda}_j^{\mathrm{E},-k})^\top\right)$, where $\boldsymbol{\Lambda}_j^{\mathrm{E}, -k}=\mathbb{E}[\boldsymbol{\Lambda}_j^{t, -k}]$. For $\forall\delta_0$, $r_1>0$ and $\delta'\in(0,1)$ such that $\mathrm{P_r}(\lvert \nu_j^{t, -k}-\Bar{\nu}_j^{-k}\rvert\geq r_1)\leq \delta'$,\footnote{The choice of Bernstein parameter $r_1$ can be obtained simply by substituting $\boldsymbol{\Lambda}_j^t$ and $\boldsymbol{\Lambda}_j^{\mathrm{E}}$ in this paper into expressions in \cite[Appendix F]{weng2025codingenforcedrobustsecureaggregation} . 
} the partial sum $\boldsymbol{S}^t_j$ is $\left(\epsilon_{j,k}^{(2)},(1-p_{j,k})\left(\delta'+\delta_j^{(2)}\right)\right)$-differentially private w.p. at least $\left(1- e^{-\frac{D}{2}(\delta_0-\ln{(1+\delta_0)})}\right)^2$ in protecting the identity of any participating node $k$, where 
\begin{align}
    \epsilon_{j,k}^{(2)}=
    \begin{cases}
        \left[2\log\left( \frac{1.25}{\delta_j^{(2)}} \right)\right]^{\frac{1}{2}}\frac{\lvert g_{j,k} \rvert\left( (\zeta+ \lambda)\sqrt{D(1+\delta_0)}\right)}{\sqrt{\Bar{\nu}_j^{-k}-r_1}}\;\; \mathrm{if}\;p_{j,k}\geq 0,\\
        \hspace{25mm}0\;\;\;\;\;\;\;\;\;\; \;\;\; \;\;   \hspace{14.5mm}\;\;\; \mathrm{if}\;p_{j,k}=0,
    \end{cases}
\end{align}
i.e., for any measurable set $\mathcal{S}$, it holds that 
\begin{align}
    \mathrm{P_r}(\boldsymbol{S}_j^{t,-k}\in \mathcal{S} )\leq e^{\epsilon_{j,k}^{(2)}}\mathrm{P_r}(\boldsymbol{S}_j^t\in \mathcal{S} )+(1-p_{j,k})(\delta'+\delta_j^{(2)}).
\end{align}
\end{theorem}
\begin{proof}[Proof Sketch]
The participation of client $k$ always lead to the simultaneous presence or absence of $\Delta\boldsymbol{\Theta}_k^t$ and $\boldsymbol{N}_k^t$, which are two Gaussian r.v.s and independent from each other. The formal proof starts with applying Lemma \ref{lemma:chi_square bound} to $\Delta\boldsymbol{\Theta}_k^t$ and $\boldsymbol{N}_k^t$, respectively, and then utilizes their independence to bound them jointly with a probability guarantee. The rest of the proofs follow similar steps in \cite{weng2025codingenforcedrobustsecureaggregation}. 
\end{proof}
Apart from the participating identity, the relay $j$ should not distinguish $\Delta\boldsymbol{\Theta}_k^t=\boldsymbol{u}$ and $\Delta\boldsymbol{\Theta}_k^{t}=\boldsymbol{u}'$ confidently.
\begin{theorem}\label{theo: relay LDP 2}
    Given secret keys $ \{\boldsymbol{N}_k^t\}_{k\in[K]}$ and the generator matrix $\boldsymbol{A}$ as in \cite[(14)]{weng2025codingenforcedrobustsecureaggregation}, its variance is $\nu_j^t=\lVert \boldsymbol{\Lambda}_j^t\boldsymbol{A}\rVert^2\boldsymbol{I}_D$. The $\Bar{\nu}_j=\mathbb{E}[\nu_j^t]$ and $\boldsymbol{\Lambda}_j^{\mathrm{E}}=\mathbb{E}[\boldsymbol{\Lambda}_j^t]$ are specified in Theorem \ref{theo: relay LDP 1}. For $\forall\delta_0$, $r_2>0$ and $\delta'\in(0,1)$ such that $\mathrm{P_r}(\lvert\nu_j^t-\Bar{\nu}_j\rvert\geq r_2)\leq \delta'$, the partial sum $\boldsymbol{S}_j^t$ is $\left(\epsilon_{j,k}^{(3)},(1-p_{j,k})\left(\delta'+\delta_j^{(3)}\right)\right)$-differentially private in protecting the perturbation of client $k$'s local data w.p. $\left(1- e^{-\frac{D}{2}(\delta_0-\ln{(1+\delta_0)})}\right)^2$, where 
    \begin{align}
    \epsilon_{j,k}^{(3)}=
    \begin{cases}
        2\cdot\left[2\log\left( \frac{1.25}{\delta_j^{(3)}} \right)\right]^{\frac{1}{2}}\frac{\lvert g_{j,k} \rvert R}{\sqrt{\Bar{\nu}_j-r_2}}\;\;\;\;\;  \mathrm{if}\;p_{j,k}\geq 0,\\
        \hspace{15mm}\;\;\;\;\;0\;\;\;\;\;\;  \hspace{15mm}\;\;\; \mathrm{if}\;p_{j,k}=0.
    \end{cases}
\end{align}
In other words, for any $\boldsymbol{u}, \boldsymbol{u'}$, and any measurable set $\mathcal{S}$, it holds that 
\begin{align*}
    \mathrm{P_r}(\boldsymbol{S}_j^t\in \mathcal{S}\vert \Delta\boldsymbol{\Theta}^t_k=\boldsymbol{u} )
    &\leq e^{\epsilon_{j,k}^{(3)}}\mathrm{P_r}(\boldsymbol{S}_j^t \in \mathcal{S}\vert \Delta\boldsymbol{\Theta}^t_k=\boldsymbol{u}')\\
    &+(1-p_{j,k})(\delta'+\delta_j^{(3)}),
    \numberthis
\end{align*}
w.p. at least $\left(1- e^{-\frac{D}{2}(\delta_0-\ln{(1+\delta_0)})}\right)^2$.
\end{theorem}
\begin{proof}[Proof Sketch]
Lemma \ref{lemma:chi_square bound} should be applied twice to characterize the ranges of $\boldsymbol{u}, \boldsymbol{u'}$. The remaining steps are similar to those in \cite{weng2025codingenforcedrobustsecureaggregation}.
\end{proof}

\subsection{Server-level LDP}
Each local model is a random variable output by the randomized training process, e.g., via data patch sampling and stochastic neural networks. When the server is curious about a specific local model $\Delta\boldsymbol{\Theta}_k^t$, $\{\Delta\boldsymbol{\Theta}_m^t\}_{m\neq k}$ can be viewed as noises that help conceal $\Delta\boldsymbol{\Theta}_k^t$. The server shall not be able to precisely infer whether a local model $\Delta\boldsymbol{\Theta}_k^t$ is included or perturbed. Recall that in H-SecCoGC, the server can either learn about the exact sum or meaningless results. Following this idea, Theorem \ref{theo: server LDP} quantifies the LDP during the successful server aggregation with a probability guarantee. 
\begin{theorem}\label{theo: server LDP}
Given $\Delta\boldsymbol{\Theta}_k^t\sim \mathcal{N}(\boldsymbol{0}, \zeta^2\boldsymbol{I}_D)$, $\forall k$, and $ \forall \delta_0>0$ such that $\mathrm{P_r}\left(\left\lVert \Delta\boldsymbol{\Theta}_k^t\right\rVert\leq \zeta\sqrt{D(1+\delta_0)}\right)\geq 1- e^{-\frac{D}{2}(\delta_0-\ln{(1+\delta_0)})}$, the global model $\boldsymbol{M}^t=\frac{1}{K}\sum_{k\in[K]}\Delta\boldsymbol{\Theta}_k^t$ is $(\epsilon^{(6)},\delta^{(6)})$-differentially
private w.p. at least $\left(1- e^{-\frac{D}{2}(\delta_0-\ln{(1+\delta_0)})}\right)^2$ in protecting the participation of client $k$, where 
    \begin{align}
        \epsilon^{(6)}=\sqrt{\frac{D(1+\delta_0)}{K-1}}\left[2\log\left( \frac{1.25}{\delta^{(6)}} \right)\right]^{\frac{1}{2}}. 
    \end{align}
In other words, for any measurable set $\mathcal{S}$, it holds that 
\begin{align}
    \mathrm{P_r}(\boldsymbol{M}^{t, -k}\in \mathcal{S} )\leq e^{\epsilon^{(6)}}\mathrm{P_r}(\boldsymbol{M}^t\in \mathcal{S} )+\delta^{(6)}.
\end{align}
Moreover, for any $\boldsymbol{u}, \boldsymbol{u'}$, it holds that 
\begin{align*}
    \mathrm{P_r}(\boldsymbol{M}^{t}&\in \mathcal{S} \vert \Delta\boldsymbol{\Theta}_k^t=\boldsymbol{u} ) \\
    &\leq  e^{\epsilon^{(7)}}\mathrm{P_r}(\boldsymbol{M}^t\in \mathcal{S} \vert \Delta\boldsymbol{\Theta}_k^t=\boldsymbol{u}')
    +\delta^{(7)}.
    \numberthis
\end{align*}
w.p. at least $\left(1- e^{-\frac{D}{2}(\delta_0-\ln{(1+\delta_0)})}\right)^2$, where
    \begin{align}
        \epsilon^{(7)}= 2\sqrt{\frac{D(1+\delta_0)}{K-1}}\left[2\log\left( \frac{1.25}{\delta^{(7)}} \right)\right]^{\frac{1}{2}}. 
    \end{align}

\end{theorem}
\begin{proof}[Proof Sketch]
The proof is similar to \cite{weng2025codingenforcedrobustsecureaggregation}.
\end{proof}

\section{Simulations}
This section evaluates the performance of the proposed scheme in comparison to the following benchmarks in relay-assisted HFL networks. 
\begin{itemize}
    \item Relay-assisted HFL \cite{9367526,9810119} with unreliable communication. In this scheme, each client is scheduled to communicate with one relay; the relay locally aggregates all the received training models from the cluster of clients and then forwards them to the server for global aggregation. This scheme provides a baseline for evaluating the impact of unreliable communication on the performance of HFL.
    \item Private relay-assisted HFL with correlated privacy noises, which can cancel out during server aggregation with full participation. The network architecture follows \cite{9367526,9810119}.
     \item Ideal HFL, where the messages are guaranteed to arrive at the server. This benchmark illustrate the best expectation of the HFL system under certain data distribution. 
     \item Our proposed scheme, H-SecCoGC, described in Section \ref{sec:H-SecCoGC}, which provides both straggler resilience and privacy preservation over the entire networks. 
\end{itemize}
\subsubsection*{System Setups}  
In the simulation, the number of clients is set to $ M = 10$. Each client is assigned an equal number of data samples. The heterogeneous data distribution is generated using the Dirichlet method, where $\Gamma=0.2$ controls the level of data imbalance across clients. A small $\Gamma$ indicates a high imbalance.  
The total number of training rounds is set to $1000$, and each client performs $I=1$ local training iteration per round. The experiments are conducted on CIFAR-10. The convolutional neural network (CNN) comprises a $(3,32)$ convolutional layer, a ReLU activation, and a max-pooling layer, followed by a $(32,32)$ convolutional layer with ReLU and max-pooling. The output is then passed through three fully connected layers with $256$, $64$, and $10$ neurons, respectively. The network is trained using Cross-entropy loss with a learning rate $0.02$. The coding parameter $s$ is set to $7$ without specification. The correlated noises are generated as Construction IV.2 in \cite{weng2025codingenforcedrobustsecureaggregation}, where the level of privacy is controlled by $\lambda$. We report the average results of multiple runs. 
These results demonstrate the robustness of H-SecCoGC against unreliable communication over the entire network, while preserving arbitrarily strong privacy. 

\subsection{Symmetric Networks}
\begin{figure}
    \centering
    \includegraphics[width=0.9\linewidth]{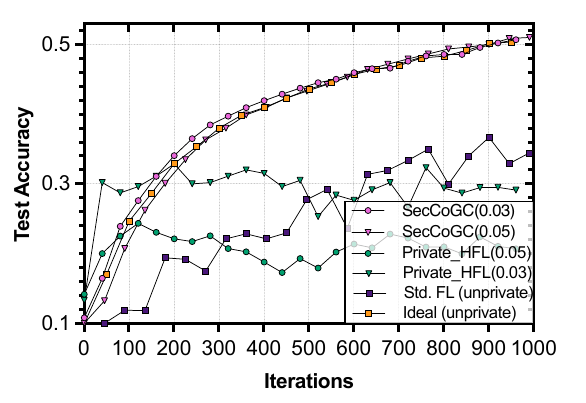}
    \vspace{-3mm}
    \caption{Test accuracy comparison of H-SecCoGC with benchmark methods under varying levels of privacy noises over unsymmetric networks. }
    \label{fig: SOTA_privacy0.05}
    \vspace{-5mm}
\end{figure}

In this setting, the client-to-relay communication probability is set to $p_{j,k}=0.9$, and the relay-to-server communication probability is set to $p_{j}=0.7$. The simulation results are presented in Fig. \ref{fig: SOTA_privacy0.05} under different privacy levels. 

\begin{figure}
    \centering
    \includegraphics[width=0.9\linewidth]{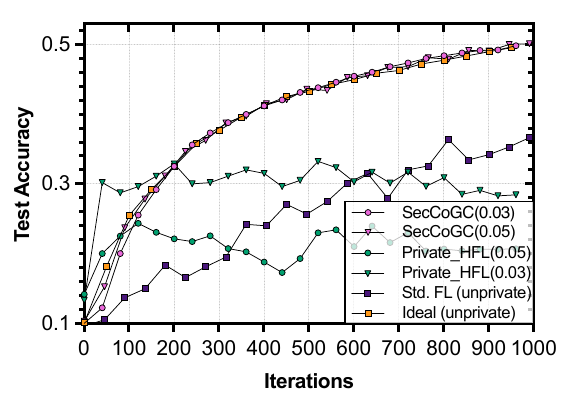}
    \vspace{-3mm}
    \caption{Test accuracy comparison of H-SecCoGC with benchmark methods under varying levels of privacy noises over symmetric networks. }
    \label{fig: SOTA_privacy_asym}
    \vspace{-5mm}
\end{figure}

It can be observed that H-SecCoGC is robust to unreliable communication and achieves optimal learning performance as the ideal case with perfect connectivity. This is attributed to the exact recovery of the global model guaranteed by the reliable coding structure supporting both aggregation of local models and cancellation of private keys. 
Compared to the unprivate HFL under symmetric but unreliable networks, H-SecCoGC enhances the reliable arrival of the local models at the server, thereby not introducing additional stochasticity, while in HFL, the global model variance, which stems from the divergence among the local models due to heterogeneous data distribution, is exaggerated by the random partial participation of clients. 
For the case of the private HFL with correlated privacy noises, when the privacy noise has relatively small power, i.e., $\lambda=0.03$, the perturbation is not too large to destroy the global model accuracy. However, when the amplitude of the privacy noises is significant compared to that of the local models, i.e., when $\lambda=0.05$, the training is prone to divergence, even if the noises can partially cancel out over intermittent relay networks due to correlation. In contrast, our proposed H-SecCoGC attains optimal performance under arbitrary privacy.

\subsection{Heterogeneous Networks}
In Fig. \ref{fig: SOTA_privacy_asym}, the superiority of H-SecCoGC is evident, consistent with the previous discussion. Under reliable communication, H-SecCoGC fully cancels the secret keys, enabling faster convergence and higher training accuracy. Moreover, its performance aligns with that of standard FL under perfect connectivity, as the coding-enforced structured aggregation effectively eliminates objective inconsistency across heterogeneous networks.

\section{Conclusion}
This paper proposes a robust secure aggregation method for HFL that ensures robust global model construction based on all local models while allowing for arbitrarily strong privacy preservation. Detailed LDP analyses are provided across all layers in H-SecCoGC, accounting for unreliable communication, stochasticity of the local models, and correlation among privacy noises. Numerical results demonstrate that the H-SecCoGC attains substantial performance improvement compared with the state-of-the-art methods.

\bibliographystyle{IEEEtran.bst}
\bibliography{IEEEabrv,reference}

\end{document}